\def\year{2018}\relax
\documentclass[letterpaper]{article} %DO NOT CHANGE THIS
\usepackage{aaai18}  %Required
\usepackage{times}  %Required
\usepackage{helvet}  %Required
\usepackage{courier}  %Required
\usepackage{url}  %Required
\usepackage{graphicx}  %Required
\usepackage{float}
\begin{document}
\title{DeepLung: 3D Deep Convolutional Nets for \\ Automated Pulmonary Nodule Detection and Classification}%2018 Formatting Instructions \\for Authors Using \LaTeX{}}
\author{
	Wentao Zhu, Chaochun Liu, Wei Fan, Xiaohui Xie \\
	University of California, Irvine \and Baidu Research
	\\
	\texttt{\{wentaoz1,xhx\}@ics.uci.edu} \and \texttt{\{liuchaochun,fanwei03\}@baidu.com}
}
%\author{Anonymous Authors\\Paper ID: 372\\
%}
\maketitle
\begin{abstract}
In this work, we present a fully automated lung CT cancer diagnosis system, DeepLung. DeepLung contains two parts, nodule detection and classification. Considering the 3D nature of lung CT data, two 3D networks are designed for the nodule detection and classification respectively. Specifically, a 3D Faster R-CNN is designed for nodule detection with a U-net-like encoder-decoder structure to effectively learn nodule features. For nodule classification, gradient boosting machine (GBM) with 3D dual path network (DPN) features is proposed. The nodule classification subnetwork is validated on a public dataset from LIDC-IDRI, on which it achieves better performance than state-of-the-art approaches, and surpasses the average performance of four experienced doctors. For the DeepLung system, candidate nodules are detected first by the nodule detection subnetwork, and nodule diagnosis is conducted by the classification subnetwork. Extensive experimental results demonstrate the DeepLung is comparable to the experienced doctors both for the nodule-level and patient-level diagnosis on the LIDC-IDRI dataset. %Code and data are available \footnote{https://github.com/***/***.git}.
\end{abstract}
\begin{figure*}
	\includegraphics[width=\linewidth]{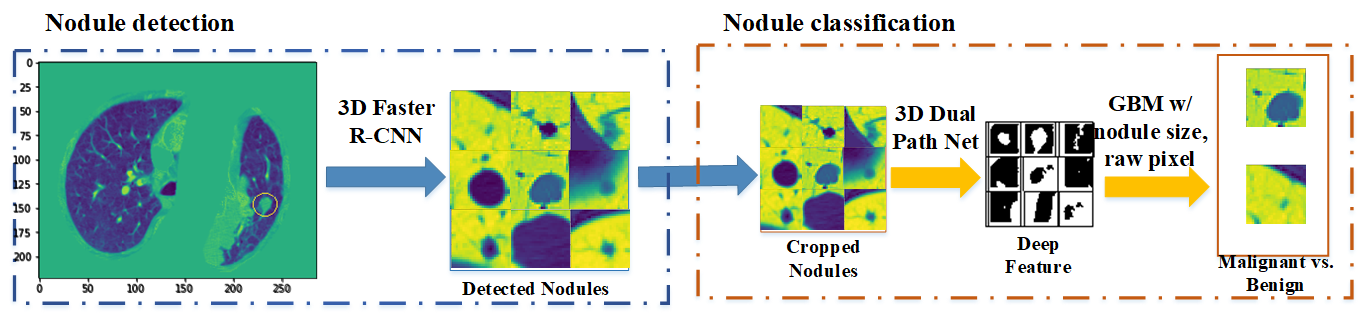}
	\caption{The framework of DeepLung. DeepLung first employs 3D Faster R-CNN to generate candidate nodules. Then it uses 3D deep dual path network (DPN) to extract deep features from the detected and cropped nodules. Lastly, gradient boosting machine (GBM) with deep features, detected nodule size, and raw nodule pixels are employed for classification. Patient-level diagnosis can be achieved by fusing the classification results of detected nodules in the CT.}
	\label{fig:framework}
\end{figure*}
\section{Introduction}
\noindent Lung cancer is the most common cause of cancer-related death in men. Low-dose lung CT screening provides an effective way for early diagnosis, which can sharply reduce the lung cancer mortality rate. Advanced computer-aided diagnosis systems (CADs) are expected to have high sensitivities while at the same time maintaining low false positive rates. Recent advances in deep learning enable us to rethink the ways of clinician lung cancer diagnosis.

Current lung CT analysis research mainly includes nodule detection \cite{dou2017automated,ding2017accurate}, and nodule classification \cite{shen2015multi,dlcls,hussein2017risk,yan2016classification}. There is few work on building a complete lung CT cancer diagnosis system for fully automated lung CT cancer diagnosis, integrating both nodule detection and nodule classification. It is worth exploring a whole lung CT cancer diagnosis system and understanding how far the performance of current technology differs from that of experienced doctors. To our best knowledge, this is the first work for a fully automated and complete lung CT cancer diagnosis system. % based on detected nodules

The emergence of large-scale dataset, LUNA16 \cite{setio2016pulmonary}, accelerates the nodule detection related research. Typically, nodule detection consists of two stages, region proposal generation and false positive reduction. Traditional approaches generally require manually designed features such as morphological features, voxel clustering and pixel thresholding \cite{murphy2009large,jacobs2014automatic}. Recently, deep convolutional networks, such as Faster R-CNN \cite{kaimingfasterrcnn} and fully convolutional networks \cite{long2015fully}, are employed to generate candidate bounding boxes \cite{ding2017accurate,dou2017automated}. In the second stage, more advanced methods or complex features, such as more carefully designed texture features, are designed to remove false positive nodules. Due to the 3D nature of CT data and the effectiveness of Faster R-CNN for object detection in 2D natural images \cite{iccv17detectorcompare}, we design a 3D Faster R-CNN for nodule detection with U-net-like encoder-decoder structure to effectively learn latent features \cite{ronneberger2015u}. %Specifically, we firstly design 3D Faster R-CNN to generate candidate nodules from raw CTs. Then we exploit multi-scale LBP features and gradient boosting machine to refine the generated bounding boxes \cite{multiscalelbp,gbt}.

Before the era of deep learning, feature engineering followed by classifiers is a general pipeline for nodule classification \cite{han2013texture}. After the public large-scale dataset, LIDC-IDRI \cite{armato2011lung}, becomes available, deep learning based methods have become dominant for nodule classification research \cite{dlcls}. Multi-scale deep convolutional network with shared weights on different scales has been proposed for the nodule classification \cite{shen2015multi}. The weight sharing scheme reduces the number of parameters and forces the multi-scale deep convolutional network to learn scale-invariant features. Inspired by the recent success of dual path network (DPN) on ImageNet \cite{chen2017dual,imagenet}, we propose a novel framework for CT nodule classification. First, we design a 3D deep dual path network to extract features. Then we use gradient boosting machines (GBM) with deep 3D dual path features, nodule size and cropped raw nodule CT pixels for the nodule classification \cite{gbt}. %In the nodule classification task, From our observation, the nodule size is an important feature for nodule classification.

%\begin{figure}
%	\includegraphics[width=\linewidth]{./fig/accwrtdim.png}
%	\caption{Diagnosis accuracy using diameter feature on fold 1. Nodule size is an important feature for nodule diagnosis.}
%	\label{fig:clsdim}
%\end{figure}  as shown in Fig. \ref{fig:clsdim}

At last, we build a fully automated lung CT cancer diagnosis system, DeepLung, by combining the nodule detection network and nodule classification network together, as illustrated in Fig. \ref{fig:framework}. For a CT image, we first use the detection subnetwork to detect candidate nodules. Next, we employ the classification subnetwork to classify the detected nodules into either malignant or benign. Finally, the patient-level diagnosis result can be achieved for the whole CT by fusing the diagnosis result of each nodule.

Our main contributions are as follows: 1) To our best knowledge, DeepLung is the first work for a complete automated lung CT cancer diagnosis system. 2) Two 3D deep convolutional networks are designed for nodule detection and classification. Specifically, inspired by the effective of Faster R-CNN for object detection and deep dual path network's success on ImageNet \cite{iccv17detectorcompare,chen2017dual}, we propose 3D Faster R-CNN for nodule detection, and 3D deep dual path network for nodule classification. 3) Our classification framework achieves superior performance compared with state-of-the-art approaches, and the performance surpasses the average performance of experienced doctors on the largest public dataset, LIDC-IDRI dataset. 4) The DeepLung system is comparable to the average performance of experienced doctors both on nodule-level and patient-level diagnosis. %3 Experimental results demonstrate the effectiveness of the detector on LUNA16 dataset. 

\section{Related Work}

Traditional nodule detection requires manually designed features or descriptors \cite{lopez2015large}. Recently, several works have been proposed to use deep convolutional networks for nodule detection to automatically learn features, which is proven to be much more effective than hand-crafted features. Setio et al. proposes multi-view convolutional network for false positive nodule reduction \cite{setio2016pulmonarymultiview}. Due to the 3D nature of CT scans, some work propose 3D convolutional networks to handle the challenge. The 3D fully convolutional network (FCN) is proposed to generate region candidates, and deep convolution network with weighted sampling is used in the false positive candidates reduction stage \cite{dou2017automated}. Ding et al. uses the Faster R-CNN to generate candidate nodules, followed by 3D convolutional networks to remove false positive nodules \cite{ding2017accurate}. Due to the effective performance of Faster R-CNN \cite{iccv17detectorcompare,kaimingfasterrcnn}, we design a novel network, 3D Faster R-CNN, for the nodule detection. Further, U-net-like encoder-decoder scheme is employed for 3D Faster R-CNN to effectively learn the features \cite{ronneberger2015u}.

Nodule classification has traditionally been based on segmentation \cite{el20113d,zhu2016adversarial} and manual feature design \cite{aerts2014decoding}. Several works designed 3D contour feature, shape feature and texture feature for CT nodule diagnosis \cite{way2006computer,el20113d,han2013texture}. Recently, deep networks have been shown to be effective for medical images. Artificial neural network was implemented for CT nodule diagnosis \cite{suzuki2005computer}. More computationally effective network, multi-scale convolutional neural network with shared weights for different scales to learn scale-invariant features, is proposed for nodule classification \cite{shen2015multi}. Deep transfer learning and multi-instance learning is used for patient-level lung CT diagnosis \cite{dlcls}. A comparison on 2D and 3D convolutional networks is conducted and shown that 3D convolutional network is superior than 2D convolutional network for 3D CT data \cite{yan2016classification}. Further, a multi-task learning and transfer learning framework is proposed for nodule diagnosis \cite{hussein2017risk,zhu2017deep}. Different from their approaches, we propose a novel classification framework for CT nodule diagnosis. Inspired by the recent success of deep dual path network (DPN) on ImageNet \cite{chen2017dual}, we design a novel totally 3D deep dual path network to extract features from raw CT nodules. Then, we employ gradient boosting machine (GBM) with the deep DPN features, nodule size, and raw nodule CT pixels for the nodule diagnosis. Patient-level diagnosis can be achieved by fusing the nodule-level diagnosis.

\section{DeepLung Framework}
The fully automated lung CT cancer diagnosis system, DeepLung, consists of two parts, nodule detection and nodule classification. We design a 3D Faster R-CNN for nodule detection, and propose gradient boosting machine with deep 3D dual path network features, raw nodule CT pixels and nodule size for nodule classification.
\subsection{3D Faster R-CNN for Nodule Detection}
\begin{figure}
	\includegraphics[width=\linewidth]{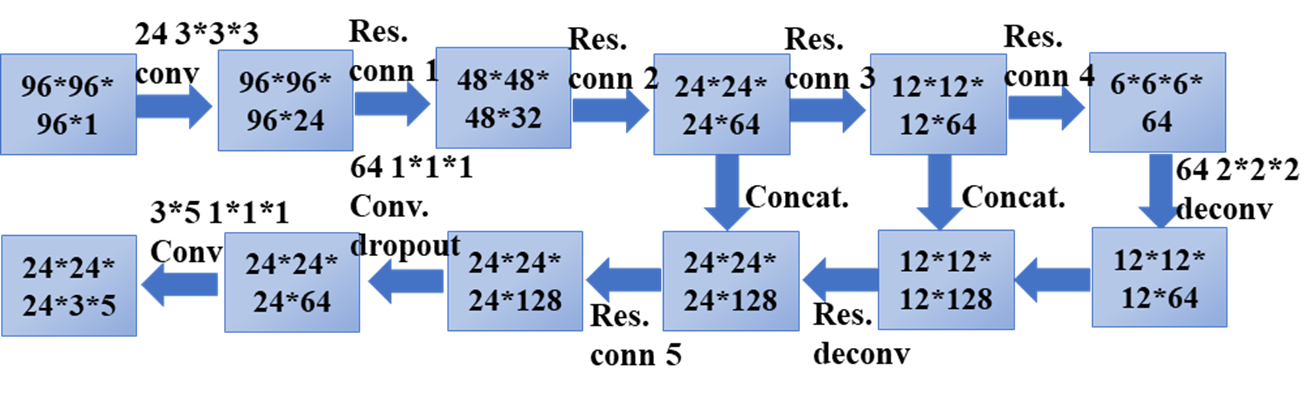}
	\caption{The 3D Faster R-CNN framework contains residual blocks and U-net-like encoder-decoder structure. The model employs 3 anchors and multi-task learning loss, including coordinates \((x, y, z)\) and diameter \(d\) regression, and candidate box classification. The numbers in boxes are feature map sizes in the format (\(\#\)slices*\(\#\)rows*\(\#\)cols*\(\#\)maps). The numbers above the connections are in the format (\(\#\)filters \(\#\)slices*\(\#\)rows*\(\#\)cols).}
	\label{fig:3dfasterrcnn}
\end{figure}

The 3D Faster R-CNN with U-net-like encoder-decoder structure is illustrated in Fig. \ref{fig:3dfasterrcnn}. Due to the GPU memory limitation, the input of 3D Faster R-CNN is cropped from 3D reconstructed CT images with pixel size \(96 \times 96 \times 96\). The encoder network is derived from ResNet-18 \cite{he2016deep}. Before the first max-pooling, two convolutional layers are used to generate features. After that, four residual blocks are employed in the encoder subnetwork. We integrate the U-net-like encoder-decoder design concept in the detection to learn the deep networks efficiently \cite{ronneberger2015u}. In fact, for the region proposal generation, the 3D Faster R-CNN conducts pixel-wise multi-scale learning and the U-net is validated as an effective way for pixel-wise labeling. This integration makes candidate nodule generation more effective. In the decoder network, the feature maps are processed by deconvolution layers and residual blocks, and are subsequently concatenated with the corresponding layers in the encoder network \cite{zeiler2010deconvolutional}. Then a convolutional layer with dropout (dropout probability 0.5) is used for the second last layer. In the last layer, we design 3 anchors, 5, 10, 20, for scale references which are designed based on the distribution of nodule sizes. For each anchor, there are 5 parts in the loss function, classification loss \(L_{cls}\) for whether the current box is a nodule or not, regression loss \(L_{reg}\) for nodule coordinates \(x, y, z\) and nodule size \(d\).

If an anchor overlaps a ground truth bounding box with the intersection over union (IoU) higher than 0.5, we consider it as a positive anchor (\(p^{\star}=1\)). On the other hand, if an anchor has IoU with all ground truth boxes less than 0.02, we consider it as a negative anchor (\(p^{\star}=0\)). The multi-task loss function for the anchor \(i\) is defined as
\begin{equation}
L(p_i, {\bf{t}_i}) = \lambda L_{cls}(p_i, p_i^{\star}) + p_i^{\star} L_{reg}({\bf{t}_i}, {\bf{t}_i}^{\star}),
\end{equation}
where \(p_i\) is the predicted probability for current anchor \(i\) being a nodule, \(\bf{t}_i\) is the regression predicted relative coordinates for nodule position, which is defined as
\begin{equation}
{\bf{t}_i} = (\frac{x-x_a}{d_a}, \frac{y-y_a}{d_a}, \frac{z-z_a}{d_a}, \log(\frac{d}{d_a})),
\end{equation}
where \((x, y, z, d)\) are the predicted nodule coordinates and diameter in the original space, \((x_a, y_a, z_a, d_a)\) are the coordinates and scale for the anchor \(i\). For ground truth nodule position, it is defined as
\begin{equation}
{{\bf{t}}_i^{\star}} = (\frac{x^{\star}-x_a}{d_a}, \frac{y^{\star}-y_a}{d_a}, \frac{z^{\star}-z_a}{d_a}, \log(\frac{d^{\star}}{d_a})),
\end{equation}
where \((x^{\star}, y^{\star}, z^{\star}, d^{\star})\) are nodule ground truth coordinates and diameter. The \(\lambda\) is set as \(0.5\). For \(L_{cls}\), we used binary cross entropy loss function. For \(L_{reg}\), we used smooth \(l_1\) regression loss function \cite{girshick2015fast}. 
\subsection{Gradient Boosting Machine with 3D Dual Path Net Feature for Nodule Classification}
\begin{figure}
	\includegraphics[width=\linewidth]{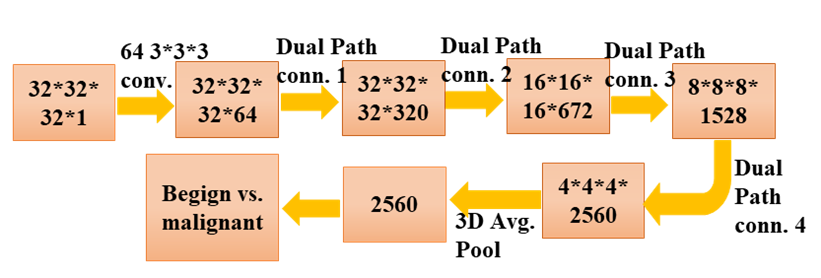}
	\caption{The 3D deep dual path network framework in the nodule classification subnetwork, which contains 4 dual path connection blocks. After the training, the deep 3D dual path network feature is extracted for gradient boosting machine to do nodule diagnosis. The numbers in the figure are of the same formats as Fig. \ref{fig:3dfasterrcnn}.}
	\label{fig:dpncls}
\end{figure}

Inspired by the success of dual path network on the ImageNet \cite{chen2017dual,imagenet}, we design a 3D deep dual path network framework for lung CT nodule classification in Fig. \ref{fig:dpncls}. % We expect this advanced network structure achieves success for CT nodule classification.

\begin{figure}
	\includegraphics[width=\linewidth]{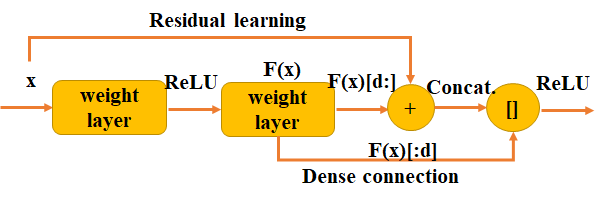}
	\caption{Illustration of dual path connection \cite{chen2017dual}, which benefits both from the advantage of residual learning \cite{he2016deep} and that of dense connection \cite{huang2016densely} from network structure design intrinsically.}
	\label{fig:dualpath}
\end{figure}

%enables feature reuse and
Dual path connection benefits both from the advantage of residual learning and that of dense connection \cite{he2016deep,huang2016densely}. The shortcut connection in residual learning is an effective way to eliminate gradient vanishing phenomenon in very deep networks. From a learned feature weights sharing perspective, residual learning enables feature reuse, while dense connection has an advantage of keeping exploiting new features \cite{chen2017dual}. And densely connected network has fewer parameters than residual learning because there is no need to relearn redundant feature maps. The assumption of dual path connection is that there might exist some redundancy in the exploited features. And dual path connection uses part of feature maps for dense connection and part of them for residual learning. In implementation, the dual path connection splits its feature maps into two parts. One part, \({\bf{F}(\bf{x})}[d:]\), is used for residual learning, the other part, \({\bf{F}(\bf{x})}[:d]\), is used for dense connection as shown in Fig. \ref{fig:dualpath}. Here \(d\) is the hyper-parameter for deciding how many new features to be exploited. The dual path connection can be formulated as
\begin{equation}
{\bf{y}} = {\bf{G}}([{\bf{F}(\bf{x})}[:d], {\bf{F}(\bf{x})}[d:]+\bf{x}]),
\end{equation}
where \(\bf{y}\) is the feature map for dual path connection, \(\bf{G}\) is used as ReLU activation function, \(\bf{F}\) is convolutional layer functions, and \(\bf{x}\) is the input of dual path connection block. Dual path connection integrates the advantages of the two advanced frameworks, residual learning for feature reuse and dense connection for keeping exploiting new features, into a unified structure, which obtains success on the ImageNet dataset.

For CT data, advanced method should be effective to extract 3D volume feature \cite{yan2016classification}. We design a 3D deep dual path network for the 3D CT lung nodule classification in Fig. \ref{fig:dpncls}. We firstly crop CT data centered at predicted nodule locations with size \(32\times 32 \times 32\). After that, a convolutional layer is used to extract features. Then 4 dual path blocks are employed to learn higher level features. Lastly, the 3D average pooling and binary logistic regression layer are used for benign or malignant diagnosis. 

The 3D deep dual path network can be used as a classifier for nodule diagnosis directly. And it can also be employed to learn effective features. 
We construct feature by concatenating the learned 3D deep dual path network features (the second last layer, 2,560 dimension), nodule size, and raw 3D cropped nodule pixels. Given complete and effective features, gradient boosting machine (GBM) is a superior method to build an advanced classifier from these features \cite{gbt}. We validate the feature combining nodule size with raw 3D cropped nodule pixels, employ GBM as a classifier, and obtain \(86.12 \%\) test accuracy averagely. Lastly, we employ GBM with the constructed feature and achieve the best diagnosis performance. % From our observation, nodule size alone provides \(82.16 \%\) test accuracy for nodule classification on average.

\subsection{DeepLung System: Fully Automated Lung CT Cancer Diagnosis}
The DeepLung system includes the nodule detection using the 3D Faster R-CNN, and nodule classification using gradient boosting machine (GBM) with constructed feature (deep dual path features, nodule size and raw nodule CT pixels) in Fig. \ref{fig:framework}. 

Due to the GPU memory limitation, we first split the whole CT into several \(96 \times 96 \times 96\) patches, process them through the detector, and combine the detected results together. We only keep the detected boxes of detection probabilities larger than 0.12 (threshold as -2 before sigmoid function). After that, non-maximum suppression (NMS) is adopted based on detection probability with the intersection over union (IoU) threshold as 0.1. Here we expect to not miss too many ground truth nodules. 

After we get the detected nodules, we crop the nodule with the center as the detected center and size as \(32 \times 32 \times 32\). The detected nodule size is kept for the classification model as a part of features. The 3D deep dual path network is employed to extract features. We use the GBM and construct features to conduct diagnosis for the detected nodules. For CT pixel feature, we use the cropped size as \(16 \times 16 \times 16\) and center as the detected nodule center in the experiments. For patient-level diagnosis, if one of the detected nodules is positive (cancer), the patient is a cancer patient, and if all the detected nodules are negative, the patient is a negative (non-cancer) patient.

\section{Experiments}
We conduct extensive experiments to validate the DeepLung system. We perform 10-fold cross validation using the detector on LUNA16 dataset. For nodule classification, we use the LIDC-IDRI annotation, and employ the LUNA16's patient-level dataset split. Finally, we also validate the whole system based on the detected nodules both on patient-level diagnosis and nodule-level diagnosis. %with pretrained model from TianChi dataset \footnote{https://tianchi.aliyun.com/competition/rankingList.htm?raceId=231601}.

In the training, for each model, we use 150 epochs in total with stochastic gradient descent optimization and momentum as 0.9. The used batch size is relied on the GPU memory. We use weight decay as \(1 \times 10^{-4}\). The initial learning rate is 0.01, 0.001 at the half of training, and 0.0001 after the epoch 120.

\subsection{Datasets}
%Tianchi dataset contains 600 training low-dose lung CTs and 200 validation low-dose lung CTs for nodule detection. The annotations are location centroids and diameters of the pulmonary nodules, which are the same with those on LUNA16 dataset \cite{setio2016pulmonary}. 

LUNA16 dataset is a subset of the largest public dataset for pulmonary nodules, the LIDC-IDRI dataset \cite{armato2011lung}. LUNA16 dataset only has the detection annotations, while LIDC-IDRI contains almost all the related information for low-dose lung CTs including several doctors' annotations on nodule sizes, locations, diagnosis results, nodule texture, nodule margin and other informations. LUNA16 dataset removes CTs with slice thickness greater than 3mm, slice spacing inconsistent or missing slices from LIDC-IDRI dataset, and explicitly gives the patient-level 10-fold cross validation split of the dataset. LUNA16 dataset contains 888 low-dose lung CTs, and LIDC-IDRI contains 1,018 low-dose lung CTs. Note that LUNA16 dataset removes the annotated nodules of size smaller than 3mm. %both Tianchi and

For nodule classification, we extract nodule annotations from LIDC-IDRI dataset, find the mapping of different doctors' nodule annotations with the LUNA16's nodule annotations, and get the ground truth of nodule diagnosis by taking different doctors' diagnosis averagely (Do not count the 0 score for diagnosis, which means N/A.). If the final average score is equal to 3 (uncertain about malignant or benign), we remove the nodule. For the nodules with score greater than 3, we label them as positive. Otherwise, we label them as negative. For doctors' annotations, we only keep those of four doctors who label most of the nodules for comparison. We only keep the CTs within LUNA16 dataset, and use the same cross validation split as LUNA16 for classification.
\subsection{Preprocessing}
We firstly clip the raw data into \([-1200, 600]\). Secondly, we transform the range linearly into \([0, 1]\). Thirdly, we use the LUNA16's given segmentation ground truth and remove the useless background. It is worth visualizing the processed data during processing because some segmentations are imperfect, and need to be removed. After that, the data can be processed by the deep networks.
\subsection{DeepLung for Nodule Detection}
We train and evaluate the detector on LUNA16 dataset following 10-fold cross validation with given patient-level split. In the training, we use the flipping, randomly scale from 0.75 to 1.25 for the cropped patches to augment the data. The evaluation metric, FROC, is the average recall rate at the average number of false positives as 0.125, 0.25, 0.5, 1, 2, 4, 8 per scan, which is the official evaluation metric for LUNA16 dataset. %Tianchi dataset and %We firstly train the detector on Tianchi dataset and obtain 77.74\% FROC on the Tianchi validation set, which is a very competitive performance.

\begin{figure}
	\begin{center}
	\includegraphics[width=0.85\linewidth]{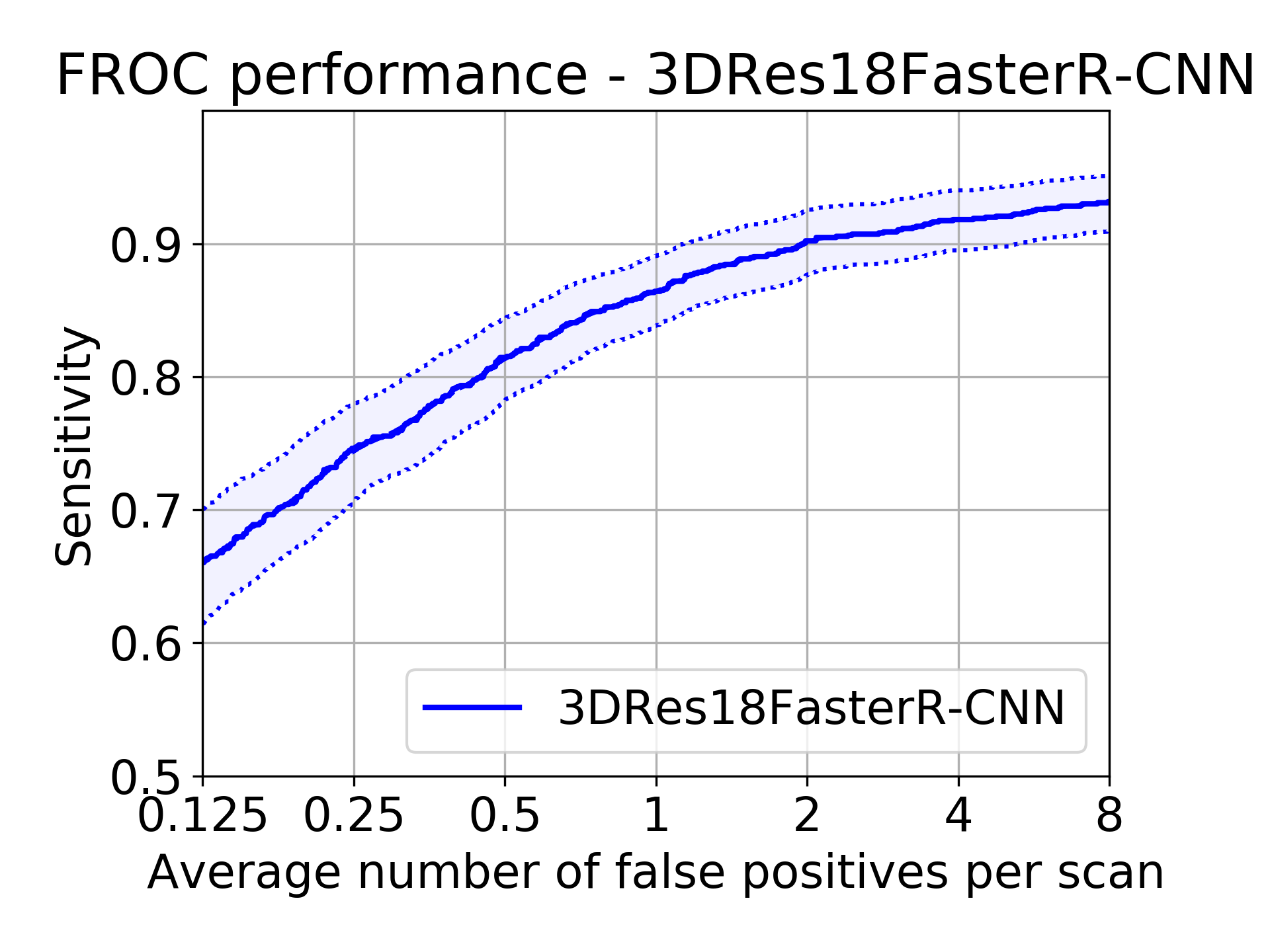}
	\caption{Sensitivity (Recall) rate with respect to false positives per scan. The FROC (average recall rate at the false positives as 0.125, 0.25, 0.5, 1, 2, 4, 8) is \textbf{83.4\%}. The 3D Faster R-CNN has a recall rate 94.6\% for all the nodules. The dash lines are lower bound and upper bound FROC for 95\% confidence interval using bootstrapping with 1,000 bootstraps \cite{setio2016pulmonary}. The solid line is the interpolated FROC based on prediction.}% If we use the top 5 returned boxes, the recall rate is 83.31\% for all the nodules.}83.41
	\label{fig:froc}
\end{center}
\end{figure}

The FROC performance on LUNA16 is visualized in Fig. \ref{fig:froc}. The solid line is interpolated FROC based on true prediction, and the dash lines are upper bound and lower bound for the bootstrapped FROC performance. The 3D Faster R-CNN achieves {\textbf{83.4\%}} FROC without any false positive nodule reduction stage, which is comparable with 83.9\% for the best method, which uses two-stage training \cite{dou2017automated}. %The recall rate is 94.62\%, which means it is fine as a detector for fully automated lung CT cancer diagnosis system because only 5\% nodules are missed.

\subsection{DeepLung for Nodule Classification}

We validate the nodule classification performance of the DeepLung system on the LIDC-IDRI dataset with the LUNA16's split principle, 10-fold patient-level cross validation. There are 1,004 nodules left and 450 nodules are positive. In the training, we firstly pad the nodules of size \(32 \times 32 \times 32\) into \(36 \times 36 \times 36\), randomly crop \(32 \times 32 \times 32\) from the padded data, horizontal flip, vertical flip, z-axis flip the data for augmentation, randomly set \(4\times4\times4\) patch as 0, and normalize the data with the mean and std. obtained from training data. The total number of epochs is 1,050. The learning rate is 0.01 at first, then became 0.001 after epoch 525, and turned into 0.0001 after epoch 840. Due to time and resource limitation for training, we use the fold 1, 2, 3, 4, 5 for test, and the final performance is the average performance on the five test folds. The nodule classification performance is concluded in Table \ref{tab:nodcls}.

\begin{table}[]
	\centering
	\caption{Nodule classification comparisons on LIDC-IDRI dataset.}
	\label{tab:nodcls}
	\begin{tabular}{c|c|c|c|c}
		\hline
		Models & \begin{tabular}[c]{@{}l@{}}3D DPN\end{tabular} & \begin{tabular}[c]{@{}l@{}}Nodule Size\\ +Pixel+GBM\end{tabular} & \begin{tabular}[c]{@{}l@{}}All feat.\\ +GBM\end{tabular} & \begin{tabular}[c]{@{}l@{}}m-CNN\end{tabular} \\ \hline
		Acc. (\%)  & 88.74                                              & 86.12                                                             & \textbf{90.44}                                            & 86.84                                                    \\ \hline
	\end{tabular}
\end{table}%. Our deep 3D dual path network feature, nodule size, and raw pixel followed by gradient boosting machine achieves the best performance.

From the table \ref{tab:nodcls}, our 3D deep dual path network (DPN) achieves better performance than that of multi-scale CNN \cite{shen2015multi} because of the power of 3D structure and deep dual path network. Because of the power of gradient boosting machine (GBM), GBM with nodule size and raw nodule pixels with crop size as \(16 \times 16 \times 16\) achieves comparable performance as multi-scale CNN \cite{shen2015multi}. Finally, we construct feature with 3D deep dual path network features, nodule size and raw nodule pixels, and obtain {\bf{90.44\%}} accuracy. % We simply use nodule size as feature for classification and obtain \(82.16 \%\) test accuracy on average. 
\subsubsection{Compared with Experienced Doctors on Their Individually Confident Nodules}

We compare our predictions with those of four experienced doctors on their individually confident nodules (with individual score not as 3). Note that about 1/3 nodules are labeled as 3 for each doctor. Comparison results are concluded in Table \ref{tab:nodclsdr}.

\begin{table}[]
	\centering
	\caption{Nodule-level diagnosis accuracy (\%) between nodule classification subnetwork in DeepLung and experienced doctors on doctor's individually confident nodules.}
	\label{tab:nodclsdr}
	\begin{tabular}{c|c|c|c|c|c}
		\hline
		& Dr 1 & Dr 2 & Dr 3 & Dr 4 & Average \\ \hline
		Doctors   & 93.44    & 93.69    & 91.82    & 86.03    & 91.25   \\ \hline
		DeepLung & 93.55    & 93.30    & 93.19    & 90.89    & \textbf{92.74}   \\ \hline
	\end{tabular}
\end{table}

From Table \ref{tab:nodclsdr}, these doctors' confident nodules are easy to be diagnosed nodules from the performance comparison between our model's performances in Table \ref{tab:nodcls} and Table \ref{tab:nodclsdr}. To our surprise, the average performance of our model is {\textbf{1.5\%}} better than that of experienced doctors. In fact, our model's performance is better than 3 out of 4 doctors (doctor 1, 3, 4) on the nodule diagnosis task. The result validates deep network surpasses human-level performance for image classification \cite{he2016deep}, and the DeepLung is better suited for nodule diagnosis than experienced doctors. 

\subsection{DeepLung for Fully Automated Lung CT Cancer Diagnosis}
We also validate the DeepLung for fully automated lung CT cancer diagnosis on the LIDC-IDRI dataset with the same protocol as LUNA16's patient-level split. Firstly, we employ the 3D Faster R-CNN to detect suspicious nodules. Then we retrain the model from nodule classification model on the detected nodules dataset. If the center of detected nodule is within the ground truth positive nodule, it is a positive nodule. Otherwise, it is a negative nodule. Through this mapping from the detected nodule and ground truth nodule, we can evaluate the performance and compare it with the performance of experienced doctors. We adopt the test fold 1, 2, 3, 4, 5 to validate the performance the same as that for nodule classification.

\begin{table}[]
	\centering
	\caption{Comparison between DeepLung's nodule classification on all detected nodules and doctors on all nodules.}
	\label{tab:nodclstpfp}
	\begin{tabular}{c|c|c|c}
		\hline
		Method& TP Set & FP Set & Doctors  \\ \hline
		Acc. (\%)   & 81.42    & 97.02    & 74.05-82.67       \\ \hline
	\end{tabular}
\end{table}

Different from pure nodule classification, the fully automated lung CT nodule diagnosis relies on nodule detection. We evaluate the performance of DeepLung on the detection true positive (TP) set and detection false positive (FP) set individually in Table \ref{tab:nodclstpfp}. If the detected nodule of center within one of ground truth nodule regions, it is in the TP set. If the detected nodule of center out of any ground truth nodule regions, it is in FP set. From Table \ref{tab:nodclstpfp}, the DeepLung system using detected nodule region obtains \(\textbf{81.42\%}\) accuracy for all the detected TP nodules. Note that the experienced doctors obtain 78.36\% accuracy for all the nodule diagnosis on average. The DeepLung system with fully automated lung CT nodule diagnosis still achieves above average performance of experienced doctors. On the FP set, our nodule classification subnetwork in the DeepLung can reduce 97.02\% FP detected nodules, which guarantees that our fully automated system is effective for the lung CT cancer diagnosis. %the DeepLung can reduce 97.02\% FP detected nodules, which shows it is effective for the fully automated lung CT cancer diagnosis scenario. %For the comparison, we split the detected nodules into detection true positive (TP) set and detection false positive (FP) set. 

\subsubsection{Compared with Experienced Doctors on Their Individually Confident CTs}

We employ the DeepLung for patient-level diagnosis further. If the current CT has one nodule that is classified as positive, the diagnosis of the CT is positive. If all the nodules are classified as negative for the CT, the diagnosis of the CT is negative. We evaluate the DeepLung on the doctors' individually confident CTs for benchmark comparison in Table \ref{tab:ctclsdr}. %because the DeepLung has about 95\% recall for all the nodules
\begin{table}[]
	\centering
	\caption{Patient-level diagnosis accuracy(\%) between DeepLung and experienced doctors on doctor's individually confident CTs.}
	\label{tab:ctclsdr}
	\begin{tabular}{c|c|c|c|c|c}
		\hline
		& Dr 1 & Dr 2 & Dr 3 & Dr 4 & Average \\ \hline
		Doctors   & 83.03    & 85.65    & 82.75    & 77.80    & \textbf{82.31}   \\ \hline
		DeepLung & 81.82    & 80.69    & 78.86    & 84.28    & 81.41   \\ \hline
	\end{tabular}
\end{table}

From Table \ref{tab:ctclsdr}, DeepLung achieves {\bf{81.41\%}} patient-level diagnosis accuracy. The performance is {\bf{99\%}} of the average performance of four experienced doctors, and the performance of DeepLung is better than that of doctor 4. Thus DeepLung can be used to help improve some doctors' performance, like that of doctor 4, which is the goal for computer aided diagnosis system. We also employ Kappa coefficient, which is a common approach to evaluate the agreement between two raters, to test the agreement between DeepLung and the ground truth patient-level labels \cite{smeeton1985early}. For comparison, we valid those of 4 individual doctors on their individual confident CTs. The Kappa coefficient of DeepLung is 63.02\%, while the average Kappa coefficient of doctors is 64.46\%. It shows the predictions of DeepLung are of good agreement with ground truths for patient-level diagnosis, and are comparable with those of experienced doctors.% The reasons for performance decrease are that, the nodule detection misses some nodules, and detection result cannot exactly match the nodule position. We will focus on the problem in our future research.
\section{Discussion}
In this section, we are trying to explain the DeepLung by visualizing the nodule detection and classification results.
\subsection{Nodule Detection}
\begin{figure}[h]
	\includegraphics[width=\linewidth]{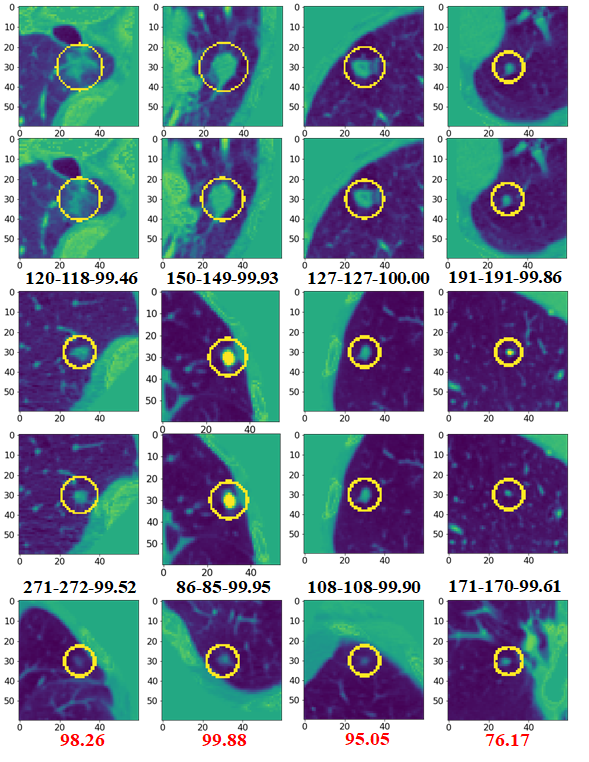}
	\caption{Visualization of central slices for nodule ground truths and detection results. We randomly choose nodules (the first row and third row) from test fold 1. Detection results are shown in the second row and fourth row. The numbers below detected nodules are with the format (ground truth slice number-detected slice number-detection probability). The last row shows the false positive detected nodules and detection probabilities. Note that the ground truth nodules are of diameter greater than 3mm. The DeepLung performs well for detection.}
	\label{fig:detection}
\end{figure}

We randomly pick nodules from test fold 1 and visualize them in the first row and third row in Fig. \ref{fig:detection}. Detected nodules are visualized in the second, fourth and fifth row. The numbers below detected nodules are with the format (ground truth slice number-detected slice number-detection probability). The fifth row with red detection probabilies are the false positive detections. Here we only visualize the slice of center z. The yellow circles are the ground truth nodules or detected nodules respectively. The center of the circle is the nodule center and the diameter of the circle is relative to the nodule size.

From the first four rows in Fig. \ref{fig:detection}, we observe the 3D Faster R-CNN works well for the nodules from test fold 1. The detection probability is also very high for these nodules. From the detected false positive nodules in the last row, we can find these false positives are suspicious to nodules visually if we observe them carefully. And the sizes of these false positive nodules are very small. Note that the nodules less than 3mm are removed in the ground truth annotations. 
\subsection{Nodule Classification}
\begin{figure}[!h]
	\includegraphics[width=\linewidth]{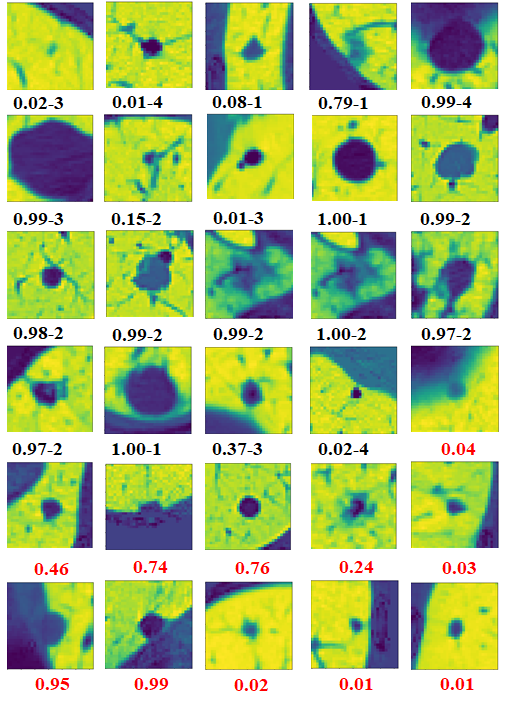}
	\caption{Visualization of central slices for nodule classification results on test fold 1. We choose nodules that is predicted right, but annotated wrong by some doctors in the first 4 rows. The numbers below the nodules are (model predicted malignant probability-which doctor is wrong). The last 2 rows show the nodules our model predicted wrong. The red number is the wrongly predicted malignant probabilities (\(>0.5\), predict malignant; \(<0.5\), predict benign). The DeepLung avoids doctors' individual bias and achieves better performance than the average performance of doctors.}
	\label{fig:classification}
\end{figure}

We also visualize the nodule classification results from test fold 1 in Fig. \ref{fig:classification}. The nodules in the first four rows are those the DeepLung predicted right, but some doctors predicted wrong. The first number is the DeepLung predicted malignant probability. The second number is which doctor predicted wrong. If the probability is large than 0.5, it is a malignant nodule. Otherwise, it is a benign nodule. The rest nodules are those the DeepLung predicted wrong with red numbers indicating the DeepLung predicted malignant probability. For an experienced doctor, if a nodule is big and has irregular shape, it has a high probability to be a malignant nodule.

From the first 4 rows in Fig. \ref{fig:classification}, we can observe that doctors mis-diagnose some nodules. The reason is that, humans are not good at processing 3D CT data, which is of low signal to noise ratio. In fact, even for high quality 2D natural image, the performance of deep network surpasses that of humans \cite{he2016deep}. From the doctors' individual experience point of view, some doctors intend to be optimistic to some nodules, and some doctors intend to be pessimistic to them. And they can just observe one slice each time. Some irregular boundaries are vague. The machine learning based methods can learn these complicated rules and high dimensional features from these doctors' annotations, and avoid such bias. %Some doctors may think 5mm diameters are big enough to be a cancer, while others think 10mm are big.

The rest nodules with red numbers are the nodules the DeepLung predicted wrong for test fold 1 in Fig. \ref{fig:classification}. If the predicted malignant probability is large than 0.5, the ground truth label is benign. Otherwise, the ground truth label is malignant. From the central slices, we can hardly decide the DeepLung's predictions are wrong. Maybe the DeepLung cannot find some weak irregular boundaries as for the bottom right 4 nodules, which is the possible reason why the DeepLung predicts them as benign. From the above analysis, the DeepLung can be considered as a tool to assist the diagnosis for doctors. Combing the DeepLung and doctor's own diagnosis is an effective way to improve diagnosis accuracy.
\section{Conclusion}
In this work, we propose a fully automated lung CT cancer diagnosis system, DeepLung. DeepLung consists two parts, nodule detection and classification. For nodule detection, we design a 3D Faster R-CNN with U-net-like encoder-decoder structure to detect suspicious nodules. Then we input the detected nodules into the nodule classification network. The 3D deep dual path network is designed to extract features. Further, gradient boosting machine with different features combined together to achieve the state-of-the-art performance, which surpasses that of experienced doctors. Extensive experimental results on public available large-scale datasets, LUNA16 and LIDC-IDRI datasets, demonstrate the superior performance of the DeepLung.

In the future, we will develop more advanced nodule detection method to further boost the performance of DeepLung. Further, a unified framework for automated lung CT cancer diagnosis is expected for nodule detection and classification. Integrating multi-modality data, such as electronic medical report history, into the DeepLung is another direction to enhance the diagnosis performance. 
\bibliographystyle{aaai}
%\bibliography{aaai18}
\small \bibliography{aaai18}

\end{document}